\documentclass[11pt,a4paper]{article}
\usepackage[hyperref]{emnlp-ijcnlp-2019}
\usepackage{times}
\usepackage{latexsym}
\usepackage{helvet}  %
\usepackage{courier}  %
\usepackage{url}
\usepackage{amsmath}
\usepackage{amsfonts}
\usepackage{graphicx}
\usepackage{color}
\usepackage{booktabs}

\aclfinalcopy %

\newcommand{\xhdr}[1]{{\noindent\bfseries #1.}}

\newcommand{\cut}[1]{}

\newcommand{\sectionrule}{\addlinespace[1ex]}

 \title{Trouble on the Horizon:\\
 Forecasting the Derailment of Online Conversations as they Develop}

\author{Jonathan P. Chang \hspace{0.3in} \\ 
  Cornell University \hspace{0.3in} \\ 
  {\tt jpc362@cornell.edu \hspace{0.1in}} \\ \And 
  Cristian Danescu-Niculescu-Mizil \\
  Cornell University \\ 
  {\tt cristian@cs.cornell.edu}}

\begin{document}

\maketitle

\begin{abstract}

Online discussions often derail into toxic exchanges between participants.  
Recent efforts mostly focused on detecting antisocial behavior after the fact, by analyzing single comments in isolation.
To provide more timely notice to human moderators, a system needs to preemptively detect that a conversation is heading towards derailment before it actually turns toxic.
This means modeling derailment as an emerging property of a conversation rather than as an isolated utterance-level event.

Forecasting 
 emerging conversational properties, however, poses several inherent modeling challenges.
First, since conversations are dynamic, a forecasting model needs to capture the flow of the discussion, rather than properties of individual comments.
Second, real conversations have an unknown horizon: they can end or derail at any time; thus a practical forecasting model needs to assess the risk in an online fashion, as the conversation develops.
In this work we introduce a conversational forecasting model that learns an unsupervised representation of conversational dynamics and exploits it to predict future derailment as the conversation develops.
By applying this model to two new diverse datasets of online conversations with labels for antisocial events, we show that it outperforms state-of-the-art systems at forecasting derailment.

\end{abstract}

\section{Introduction}
\label{sec:intro}
\begin{quote}
\small
\vspace{-0.01cm}
``Ch\'e saetta previsa vien pi\`u lenta.''\footnote{``The arrow one foresees arrives more gently.''}
\end{quote}
 \vspace{-0.34cm}
\begin{flushright}
\small
-- Dante Alighieri, Divina Commedia, Paradiso
\end{flushright}

\begin{figure}
    \centering
    \footnotesize
    \begin{tabular}{p{0.925\linewidth}}
        \toprule
        (1) \textbf{[User A]} What does [quote omitted] refer to? I assume it should be written from June 2010 to December 2011 and we should precise [sic] the months. \\
        (2) \textbf{[User B]} No. It refers to 2007-2011 Belgian political crisis \\
        (3) \textbf{[User A]} 2007-2011 Belgian political crisis is a little bit 
        [of original research].
         It merges 2 crisis in 1. Sources 2 and 4 in the article talk about a 18 months crisis in 2010-2011, ie what I refer to. What are the reliable sources that make this crisis go back to 2007? \\
        (4) \textbf{[User B]} Yes it's not ridiculous at all to claim [it's original research] because it doesn't fit your argument. A crisis can be composed out of several smaller crisis. It's not original research if some of your sources only talk about parts [...] \\
        (5) \textbf{[User A]} Where is the source that claim the crisis is 4 year long? Sources state claim it is 18 month long and refer to the period from June 2010 to December 2011. \\
        (6) \textbf{[User B]} There were 4 governments and 2 years of no government in 4 years time. You can not sanely claim that this Must be viewed as two seperate crisis. What exactly splits them up? [...] \\
        \bottomrule
    \end{tabular}
            \caption{Example start of a conversation that will eventually derail into a personal attack.
            }
        \label{fig:intro}
\end{figure}

Antisocial behavior is a persistent problem plaguing online conversation platforms; it is both widespread \cite{duggan_online_2014} and potentially damaging to mental and emotional health \cite{raskauskas_involvement_2007, akbulut_cyberbullying_2010}.
The strain 
this phenomenon
puts on community maintainers 
has
sparked recent interest in computational approaches for assisting human moderators.

Prior work in this direction has largely focused
on \emph{post-hoc} identification of various kinds of antisocial behavior, including hate speech \cite{warner_detecting_2012, davidson_automated_2017}, harassment \cite{yin_detection_2009}, personal attacks \cite{wulczyn_ex_2017}, and general toxicity \cite{pavlopoulos_deeper_2017}.  The fact that these approaches only identify antisocial content \emph{after the fact} limits their practicality as tools for assisting \emph{pre-emptive} moderation in conversational domains.

Addressing this limitation requires \emph{forecasting} the future derailment of a conversation based on early warning signs, giving the moderators time to potentially intervene before any harm is done 
(\citeauthor{liu_forecasting_2018}~\citeyear{liu_forecasting_2018}, \citeauthor{zhang_conversations_2018}~\citeyear{zhang_conversations_2018}, see  \citeauthor{jurgens_just_2019} \citeyear{jurgens_just_2019} for a discussion). Such a goal recognizes derailment as emerging from the development of the conversation, and belongs to the broader area of \emph{conversational forecasting},
which includes future-prediction tasks such as 
predicting the eventual length of a conversation \cite{backstrom_characterizing_2013},
whether a persuasion attempt will 
eventually succeed
 \cite{tan_winning_2016,wachsmuth_retrieval_2018,yang_lets_2019}, whether team discussions will 
 eventually
 lead to an increase in performance \cite{niculae_conversational_2016}, or whether ongoing counseling conversations will 
 eventually be perceived as helpful \cite{althoff_large-scale_2016}.\footnote{We can distinguish two types of forecasting tasks, depending on whether the to-be-forecasted target is an event that might take place within the conversation (e.g., derailment) or an outcome measured after the conversation will eventually conclude (e.g.,  helpfulness).  The following discussion of modeling challenges holds for both. 
 }

Approaching such \emph{conversational forecasting} problems, however, requires overcoming several inherent modeling challenges.  
First, conversations are \emph{dynamic} and their outcome might depend on how subsequent comments interact with each other.  
Consider the example in Figure \ref{fig:intro}: 
while no individual comment is outright offensive,
a human reader can sense a tension emerging from their succession 
(e.g., dismissive answers to repeated questioning).
Thus a forecasting model needs to capture not only the content of each 
individual comment,
but also the 
\emph{relations}
between comments.  
Previous work has largely relied on 
hand-crafted
 features to capture such relations---e.g., similarity between comments \cite{althoff_large-scale_2016,tan_winning_2016} or conversation structure \cite{zhang_characterizing_2018,hessel_somethings_2019}---,
though neural attention architectures have also recently shown promise
\cite{jo_attentive_2018}.

The second modeling challenge stems from the fact that conversations have an \emph{unknown horizon}: they can be of varying lengths, and the to-be-forecasted event can occur at any time.  
So when 
is it a good time to make a forecast?
Prior work has largely proposed two solutions, both resulting in important practical limitations. 
One solution is to assume (unrealistic) prior knowledge of when the to-be-forecasted event takes place and extract features up to that point \cite{niculae_linguistic_2015,liu_forecasting_2018}.  
Another compromising solution is to extract features from a fixed-length window, often at the start of the conversation \cite[inter alia]{curhan_thin_2007,niculae_conversational_2016,althoff_large-scale_2016,zhang_conversations_2018}. Choosing a catch-all window-size is however impractical: short windows will miss information in comments they do not encompass (e.g., a window of only two comments would miss the 
chain of repeated questioning
in comments 3 through 6 of Figure \ref{fig:intro}), while longer windows 
risk 
missing
 the to-be-forecasted event 
altogether
 if it 
 occurs
  before the end of the window, which would prevent \emph{early} detection.

In this work we introduce a 
model for forecasting conversational events 
that overcomes both these inherent challenges by processing comments, and their
relations,
as they happen (i.e., in an online fashion).  
Our main insight is that models with these properties already exist,
albeit geared toward generation rather than prediction:
recent work in context-aware dialog generation (or ``chatbots'') has proposed sequential neural models that make effective use of the intra-conversational dynamics \cite{sordoni_neural_2015, serban_building_2016, serban_hierarchical_2017}, while concomitantly being able to process the conversation 
as it develops
(see \citet{gao_neural_2018} for a survey).  

In order for these systems to perform well in the generative domain they need to be trained on massive amounts of (unlabeled) conversational data.  
The main difficulty in directly adapting these models to the supervised domain of conversational forecasting is the relative scarcity of labeled data: for most forecasting tasks, at most a few thousands labeled examples are available, insufficient for the notoriously data-hungry sequential neural models.

To overcome this difficulty, we propose to decouple the objective of learning a neural representation 
of conversational dynamics from the objective of predicting future events.
The former can be \emph{pre-trained} on large amounts of unsupervised data, similarly 
to how chatbots are trained. 
The latter can piggy-back on the resulting representation after \emph{fine-tuning} 
it for classification using relatively small labeled data.
While similar pre-train-then-fine-tune approaches have recently achieved state-of-the-art performance in a number of NLP tasks---including natural language inference, question answering, and commonsense reasoning (discussed in Section \ref{sec:related})---to the best of our knowledge this is the first attempt at applying this paradigm to conversational forecasting.

To test the effectiveness of this new architecture in forecasting derailment of online conversations, we develop and distribute two new datasets.  
The first 
triples in size
the highly curated 
`Conversations Gone Awry' dataset \cite{zhang_conversations_2018},
where civil-starting Wikipedia Talk Page conversations are crowd-labeled according to whether they eventually lead to personal attacks;
the second relies on in-the-wild moderation of the popular subreddit ChangeMyView, where the aim is to forecast whether a discussion will 
later be subject to moderator action due to ``rude or hostile'' behavior. 
In both datasets, our model outperforms 
existing
 fixed-window approaches, as well as simpler sequential baselines that 
cannot account for inter-comment relations.  
Furthermore, by virtue of its online processing of the conversation, our system can provide substantial prior notice of upcoming derailment, triggering on average 3 comments (or 3 hours) before an overtly toxic comment is posted.

To summarize, in this work we:
\begin{itemize}
    \item introduce the first model for 
    forecasting conversational events
    that can capture the dynamics of a conversation \emph{as it develops};
    \item build two diverse datasets 
    (one entirely new, one extending prior work) 
    for the task of forecasting derailment of online conversations;
    \item compare the performance of our model 
    against the current state-of-the-art,
     and evaluate its ability to provide \emph{early} warning signs. 
\end{itemize}
Our work is motivated by the goal of assisting human moderators of online communities by preemptively signaling at-risk conversations that might deserve 
their
attention. However, we caution that any automated systems might encode or even amplify the biases existing in the training data \cite{park_reducing_2018,sap_risk_2019,wiegand_detection_2019}, so a public-facing implementation would need to be exhaustively  scrutinized for such biases \cite{feldman_certifying_2015}.

\vspace{0.04in}
\section{Further Related Work}
\vspace{0.03in}
\label{sec:related}
\xhdr{Antisocial behavior}
Antisocial behavior online comes in many forms, including harassment \cite{vitak_identifying_2017}, cyberbullying \cite{singh_they_2017}, and general aggression \cite{kayany_contexts_1998}. 
Prior work
has sought to understand different aspects of such behavior,
including its effect on the communities where it happens \cite{collier_conflict_2012, arazy_stay_2013}, the actors involved \cite{cheng_anyone_2017, volkova_identifying_2017, kumar_community_2018, ribeiro_characterizing_2018} and connections to the outside world \cite{olteanu_effect_2018}.

\xhdr{Post-hoc classification of conversations}
There is a rich body of prior work on classifying the outcome
 of a conversation after 
  it has concluded, or classifying conversational events after they happened. 
Many examples exist, but some 
more
 closely related to our present work include identifying the winner of a debate \cite{zhang_conversational_2016,potash_towards_2017,wang_winning_2017}, identifying successful negotiations \cite{curhan_thin_2007,cadilhac_grounding_2013}, as well as detecting whether deception \cite{girlea_psycholinguistic_2016,perez-rosas_verbal_2016,levitan_linguistic_2018} or disagreement \cite{galley_identifying_2004,abbott_how_2011,allen_detecting_2014,wang_piece_2014,rosenthal_i_2015} has occurred.

Our goal is different because we wish to \emph{forecast} conversational events before they happen and while the conversation is still ongoing (potentially allowing for interventions).
Note that some 
post-hoc tasks can also be re-framed as forecasting tasks (assuming the existence of necessary labels); for instance, predicting whether an ongoing conversation \emph{will} eventually spark disagreement \cite{hessel_somethings_2019}, rather than detecting already-existing disagreement.

\xhdr{Conversational forecasting}
As described in Section \ref{sec:intro}, prior work on 
 forecasting conversational outcomes and events
 has largely relied on 
hand-crafted
 features to capture aspects of conversational dynamics.
Example feature sets include statistical measures based on  
similarity between utterances \cite{althoff_large-scale_2016},
 sentiment imbalance \cite{niculae_linguistic_2015}, flow of ideas \cite{niculae_linguistic_2015}, 
 increase in hostility \cite{liu_forecasting_2018},
 reply rate \cite{backstrom_characterizing_2013}
  and graph representations of conversations \cite{garimella_quantifying_2017,zhang_characterizing_2018}.
By contrast, 
  we aim to automatically learn neural representations of 
conversational
 dynamics through pre-training.

Such 
hand-crafted
features are typically extracted from fixed-length windows of the conversation, leaving unaddressed the problem of unknown horizon.  While some work has trained \emph{multiple} models for different window-lengths \cite{liu_forecasting_2018,hessel_somethings_2019}, they consider these models to be independent and, as such, do not address the issue of aggregating them into a single forecast (i.e., deciding at what point to make a prediction). 
We implement a 
simple
sliding
windows 
solution as a baseline (Section \ref{sec:eval}).

\xhdr{Pre-training for NLP}
The use of pre-training for natural language tasks has been growing in popularity after recent breakthroughs demonstrating improved performance on a wide array of benchmark tasks \cite{peters_deep_2018,radford_improving_2018}.
Existing work has generally used a language modeling objective as the pre-training objective; examples include next-word prediction \cite{howard_universal_2018}, sentence autoencoding, \cite{dai_semi-supervised_2015}, and machine translation \cite{mccann_learned_2017}.
BERT \cite{devlin_bert_2019} introduces a variation on this in which the goal is to predict the next sentence in a document given the current sentence.
Our pre-training objective is similar in spirit, but operates at a \emph{conversation} level, 
rather than a document level.
We hence view our objective as \emph{conversational modeling} rather than (only) language modeling.
Furthermore, while BERT's sentence prediction objective is framed as a multiple-choice task, our objective is framed as a generative task.

\vspace{0.1in}
\section{Derailment Datasets}
\vspace{0.05in}
\label{sec:data}

We consider two datasets, representing related but slightly different forecasting tasks.
The first dataset is an expanded version of the annotated Wikipedia conversations dataset from \citet{zhang_conversations_2018}.
This dataset uses carefully-controlled crowdsourced labels, strictly filtered to 
ensure the conversations are civil up to the moment of a personal attack.
This is a useful property for the purposes of model analysis, and hence we focus on this as our primary dataset.
However, we are conscious of the possibility that these strict labels may not fully capture the kind of behavior that moderators care about in practice.
We therefore introduce a secondary dataset, constructed from the subreddit ChangeMyView (CMV) that does not use post-hoc annotations.
Instead, the prediction task is 
to forecast whether the conversation will be subject to moderator action in the future.

\xhdr{Wikipedia data}
\citeauthor{zhang_conversations_2018}'s `Conversations Gone Awry' dataset consists of 1,270 conversations that took place between Wikipedia editors on publicly accessible talk pages.
The conversations are sourced from the 
WikiConv
 dataset \cite{hua_wikiconv_2018} and labeled by crowdworkers as either containing a \emph{personal attack} from within (i.e., hostile behavior by one user in the conversation directed towards another) or remaining civil throughout.

A series of controls are implemented to prevent models from picking up on trivial correlations.
To prevent models from capturing topic-specific information (e.g., political conversations are more likely to derail), each attack-containing conversation is paired with a clean conversation from the same talk page, where the talk page serves as a proxy for topic.\footnote{Paired conversations were also enforced to be similar in length, so that length distribution is the same between classes.}
To force models to actually capture conversational dynamics rather than 
detecting already-existing toxicity, human annotations are used to ensure that all comments preceding a personal attack are civil.

To the ends of more effective model training, we elected to expand the 
`Conversations Gone Awry'
 dataset, using 
the original annotation procedure.
Since 
we found that
 the original data skewed 
 towards shorter conversations, we focused this crowdsourcing run on longer conversations: ones with 4 or more comments preceding the attack.\footnote{We cap the length at 10 to avoid overwhelming the crowdworkers.}
Through this additional crowdsourcing, we 
expand
 the 
 dataset to 4,188 conversations,
   which we are publicly releasing as part of the Cornell Conversational Analysis Toolkit (ConvoKit).\footnote{\href{http://convokit.cornell.edu}{convokit.cornell.edu}}

We perform an 80-20-20 train/dev/test split,
 ensuring that paired conversations end up in the same split in order to preserve the topic control.
Finally, we randomly sample another 1 million conversations from 
WikiConv
 to use for the unsupervised pre-training of the generative component.

\xhdr{Reddit CMV data}
The CMV dataset is constructed from conversations collected via the Reddit API.
In contrast to the 
Wikipedia-based
 dataset, we explicitly avoid the use of post-hoc annotation.
Instead, we use as our label whether a conversation eventually had a comment removed by a moderator for violation of Rule 2: ``Don't be rude or hostile to other users''.\footnote{The existence of this specific rule, the standardized moderation messages and the civil character of the ChangeMyView subreddit was our initial motivation for choosing it.}

Though the lack of post-hoc annotation limits the degree to which we can impose controls on the data 
(e.g., some conversations may contain toxic comments not flagged by the moderators)
we do reproduce as many of the Wikipedia data's controls as we can.
Namely, we replicate the topic control pairing by choosing pairs of positive and negative examples that 
belong to the same top-level post, following \citet{tan_winning_2016};\footnote{The top-level post is not part of the conversations.} 
and enforce that the removed comment was made by a user who was previously involved in the conversation.\footnote{We 
also
impose the same length restriction on the number of comments preceding the removed comment, for comparability and for computational considerations.}
This process results in 6,842 conversations, to which we again apply a pair-preserving 80-20-20 split. Finally, we gather over 600,000 conversations that do not include any removed comment, for unsupervised pre-training.

\section{Conversational Forecasting Model}
\label{sec:method}

\begin{figure}
    \centering
    \includegraphics[width=0.99\linewidth]{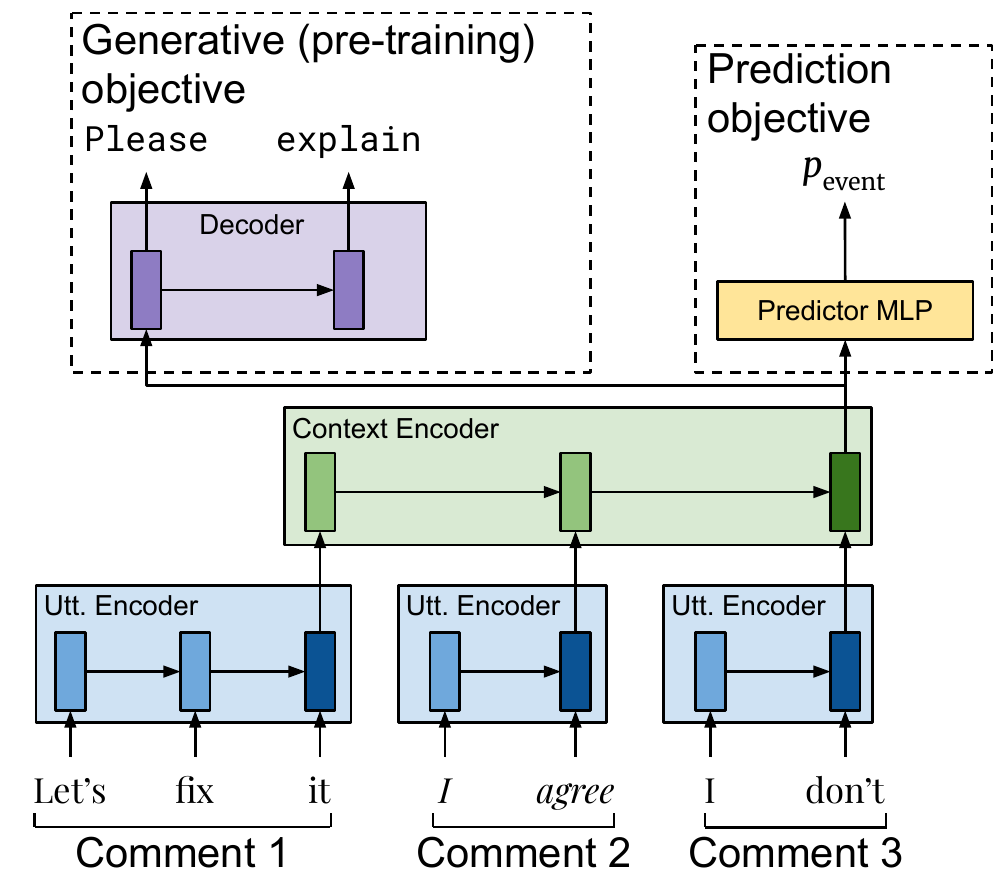}
    \caption{Sketch of the CRAFT architecture.
    }
    \label{fig:network}
\end{figure}

We now 
describe our 
general
model for forecasting future conversational 
events. 
Our model integrates two components: (a) a generative dialog model that learns to represent conversational dynamics in an unsupervised fashion; and (b) a supervised component that fine-tunes this representation to forecast future 
events. 
Figure \ref{fig:network} provides an overview of the proposed architecture, henceforth  
CRAFT (Conversational Recurrent Architecture for ForecasTing).

\xhdr{Terminology}
For modeling purposes, we treat a conversation as a sequence of $N$ comments $C = \{c_1,\dots,c_N\}$. Each comment, in turn, is a sequence of tokens, where the number of tokens may vary from comment to comment. For the $n$-th comment ($1 \le n \le N)$, we let $M_n$ denote the number of tokens. Then, a 
comment
 $c_n$ can be represented as a sequence of $M_n$ tokens: $c_n = \{w_1,\dots,w_{M_n}\}$.

\xhdr{Generative component}
For the generative component of our model, we use a hierarchical recurrent encoder-decoder (HRED) architecture \cite{sordoni_hierarchical_2015}, a modified version of the popular sequence-to-sequence (seq2seq) architecture \cite{sutskever_sequence_2014} designed to account for dependencies between consecutive inputs.
\citet{serban_building_2016}  showed that HRED can successfully model conversational context by encoding the temporal structure of previously seen comments, making it an ideal fit for our use case.
Here, we provide a high-level summary of the HRED architecture, deferring deeper technical discussion to \citet{sordoni_hierarchical_2015} and \citet{serban_building_2016}.

An HRED dialog model consists of three components: an {utterance encoder}, a {context encoder}, and a {decoder}.
The utterance encoder is responsible for generating semantic vector representations of comments.
It consists of a recurrent neural network (RNN) that reads a comment token-by-token, and on each token $w_m$ updates a hidden state $h^{\text{enc}}$ based on the current token and the previous hidden state:
\begin{equation}
    h^{\text{enc}}_m = f^{\text{RNN}}(h^{\text{enc}}_{m-1}, w_m)
    \label{eq:encoder}
\end{equation}
where $f^{\text{RNN}}$ is a nonlinear gating function 
(our implementation uses GRU \cite{cho_learning_2014}).
The final hidden state $h^{\text{enc}}_M$ can be viewed as a vector encoding of the entire comment.

Running the encoder on each comment $c_n$ results in a sequence of $N$ vector encodings.
A second encoder, the context encoder, is then run over this sequence:
\begin{equation}
    h^{\text{con}}_n = f^{\text{RNN}}(h^{\text{con}}_{n-1}, h^{\text{enc}}_{M_n})
    \label{eq:context_enc}
\end{equation}

Each hidden state $h^{\text{con}}_n$ can then be viewed as an encoding of the full conversational context up to and including the $n$-th comment.
To generate a response to comment $n$, the context encoding $h^{\text{con}}_n$ is used to initialize the hidden state 
$h^{\text{dec}}_{0}$
of a decoder RNN.
The decoder produces a response token by token using the following recurrence:
\begin{equation}
    \begin{array}{rcl}
    h^{\text{dec}}_t & = & f^{\text{RNN}}(h^{\text{dec}}_{t-1}, w_{t-1}) \\
    w_t & = & f^{\text{out}}(h^{\text{dec}}_t)
    \end{array}
    \label{eq:decoder}
\end{equation}
where $f^{\text{out}}$ is some function that outputs a probability distribution over words; we implement this using a simple feedforward layer.
In our implementation, we further augment the decoder with attention \cite{bahdanau_neural_2014,luong_effective_2015} over context encoder states to help capture long-term inter-comment dependencies.
This generative component can be pre-trained using unlabeled conversational data.

\xhdr{Prediction component}
Given a pre-trained HRED dialog model, we aim to extend the model to predict from the conversational context whether the 
to-be-forecasted 
event
 will occur.
Our predictor consists of a multilayer perceptron (MLP) with 3 fully-connected layers, leaky ReLU activations between layers, and sigmoid activation for output.
For each comment $c_n$, the predictor takes as input the context encoding $h^{\text{con}}_n$ and forwards it through the MLP layers, resulting in an output score that is interpreted as a probability $p_{\text{event}}(c_{n+1})$ that the to-be-forecasted 
event
will happen (e.g., that the conversation will derail).

Training the predictive component starts by initializing the weights of the encoders to the values learned in pre-training.
The main training loop then works as follows: 
for each positive sample---i.e., a conversation containing an instance of the to-be-forecasted event (e.g., derailment) at comment $c_e$---we feed the context $c_1,\dots,c_{e-1}$ through the encoder and classifier, and compute 
cross-entropy loss between the classifier output and expected output of 1.
Similarly, for each negative sample---i.e., a conversation where none of the comments exhibit the to-be-forecasted event and that ends with $c_N$---we feed the context $c_1,\dots,c_{N-1}$ through the model and compute loss against an expected output of 0.

Note that the parameters of the generative component are not held fixed during this process; instead, backpropagation is allowed to go all the way through the encoder layers.
This process, known as \emph{fine-tuning}, reshapes the representation learned during pre-training to be more directly useful to prediction \cite{howard_universal_2018}.

We implement the model and training code using PyTorch, and we are publicly releasing our implementation and the trained models together with the data 
as part of ConvoKit.

\begin{table*}
    \centering
    \begin{tabular}{rcccccccc|ccccc}
        \toprule
        & \multicolumn{3}{c}{Capabilities} &  \multicolumn{5}{c}{Wikipedia Talk Pages} & \multicolumn{5}{c}{Reddit CMV} \\
        \textbf{Model} & \textbf{D} & \textbf{O} & \textbf{L} & \textbf{A} & \textbf{P} & \textbf{R} & \textbf{FPR} & \textbf{F1} & \textbf{A} & \textbf{P} & \textbf{R} & \textbf{FPR} & \textbf{F1} \\
        \midrule
        BoW & & & & 56.5 & 55.6 & 65.5 & 52.4 & 60.1 & 52.1 & 51.8 & 61.3 &  57.0 & 56.1 \\
        Awry & \checkmark & & &  58.9 & 59.2 & 57.6 & 39.8 & 58.4 & 54.4 & 55.0 & 48.3 &  \textbf{39.5} & 51.4 \\
        \sectionrule
        Cumul. BoW &  & \checkmark & &  60.6 & 57.7 & \textbf{79.3} & 58.1 & 66.8 & 59.9 & 58.8 & 65.9 & 46.2 & 62.1 \\
        Sliding Awry & \checkmark & \checkmark & & 60.6 & 60.2 & 62.4 & 41.2 & 61.3 & 56.8 & 56.6 & 58.2 & 44.6 & 57.4 \\
        \sectionrule
        CRAFT $-$ CE & & \checkmark & \checkmark & 64.9 & \textbf{64.4} & 66.7 & \textbf{36.9} & 65.5  & 57.7 & 56.1 & 71.2 & 55.7 & 62.8 \\
        CRAFT & \checkmark & \checkmark & \checkmark & \textbf{66.5} & 63.7 & 77.1 & 44.1 & \textbf{69.8} & \textbf{63.4} & \textbf{60.4} & \textbf{77.5} & 50.7 & \textbf{67.9} \\
        \midrule
    \end{tabular}
    \caption{Comparison of the capabilities of each baseline and our CRAFT models (full and without the Context Encoder) with regards to capturing inter-comment (D)ynamics, processing conversations in an (O)nline fashion, and automatically (L)earning feature representations, as well as their performance in terms of (A)ccuracy, (P)recision, (R)ecall, False Positive Rate~(FPR), and F1 score. Awry is the 
     model previously proposed by \citet{zhang_conversations_2018} for this task. }
    \label{tab:results}
\end{table*}

\vspace{0.1in}
\section{Forecasting Derailment}
\vspace{0.05in}
\label{sec:eval}

We evaluate the performance of CRAFT in the task of forecasting conversational derailment in both the Wikipedia and CMV scenarios.  To this end, for each of these datasets we pre-train the generative component
on the unlabeled portion of the data and fine-tune it on the labeled training split (data size detailed in Section \ref{sec:data}).

In order to evaluate our sequential system against conversational-level ground truth, we need to aggregate comment level predictions.  
If \emph{any} comment in the conversation \emph{triggers} a positive prediction---i.e., $p_{\text{event}}(c_{n+1})$ is greater than a threshold learned on 
the development split---then the respective conversation is predicted to derail. 
If this forecast is triggered in a conversation that actually derails, but \emph{before}
the derailment actually happens, then the conversation is counted as a true positive; otherwise it is a false positive.
If no positive predictions are triggered for a conversation, but it actually derails then it counts as a false negative; if it does not derail then it is a true negative.
\xhdr{Fixed-length window baselines}
We first seek to compare CRAFT to existing, fixed-length window approaches to forecasting.
To this end, we 
implement
 two 
 such
  baselines: \emph{Awry}, 
which is the state-of-the-art method proposed in 
\citet{zhang_conversations_2018} based on pragmatic features in the first comment-reply pair,\footnote{We use the ConvoKit implementation.} 
and \emph{BoW}, a simple bag-of-words baseline that makes a prediction using TF-IDF weighted bag-of-words features extracted from the first comment-reply pair.

\xhdr{Online forecasting baselines}
Next, we consider
simpler approaches for making forecasts as the conversations happen 
(i.e., in an online fashion).
First, we propose \emph{Cumulative BoW}, a model that recomputes bag-of-words features on all comments seen thus far every time a new comment arrives.
While this approach does exhibit the desired behavior of producing updated predictions for each new comment, it fails to account for relationships between comments.

This simple cumulative approach cannot be directly extended to models whose features are strictly based on a fixed number of comments, like Awry. An alternative is to use a \emph{sliding window}: 
for a feature set based on a window of $W$ comments, upon each new comment we can extract features from a window containing that comment and the $W-1$ comments preceding it.  
We apply this to the Awry method and call this model \emph{Sliding Awry}.
For both these baselines, we aggregate comment-level predictions in the same way as in our main model.

\xhdr{CRAFT ablations}
Finally, we consider 
two modified versions of the CRAFT model in order to evaluate the impact of two of its key components: (1) the pre-training step, and (2)  
its ability to capture inter-comment dependencies through its hierarchical memory.

To evaluate the impact of pre-training, we train the 
prediction component of CRAFT
on only the labeled training data, without first pre-training the encoder layers with the unlabeled data.
We find that given the relatively small size of labeled data, this baseline fails to successfully learn, and ends up performing at the level of random guessing.\footnote{We thus exclude this baseline from the results summary.}
This result underscores the need for the pre-training step that can make use of unlabeled data.

To evaluate the impact of the 
 hierarchical memory, we implement a simplified version of CRAFT where the 
memory 
size of the context encoder is zero (\emph{CRAFT $-$ CE}), thus effectively acting as if the pre-training component is a vanilla seq2seq model.
In other words, this model cannot capture inter-comment dependencies, and instead at each step makes a prediction based only on the utterance encoding of the 
latest
comment.

\begin{figure}
    \centering
    \includegraphics[width=\linewidth]{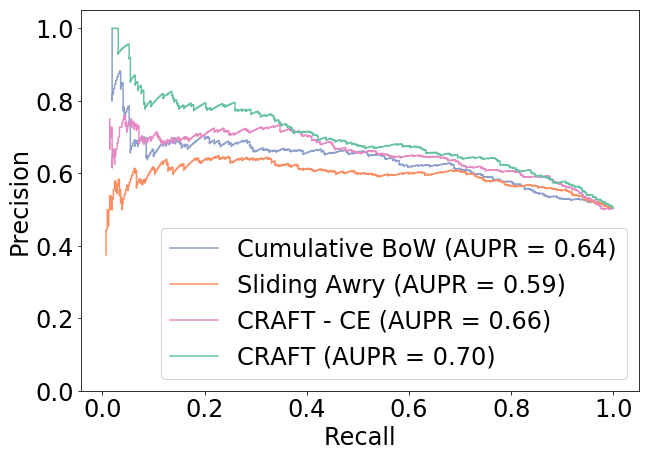}
    \caption{Precision-recall curves 
    and the area under each curve.
     To reduce clutter, we show only the curves for Wikipedia data
     (CMV curves are similar)
      and exclude the fixed-length window baselines (which perform worse).
      \vspace{0.1in}
      }
    \label{fig:pr_curve}
\end{figure}

\xhdr{Results}
Table \ref{tab:results} compares CRAFT to the baselines on the test splits (random baseline is 50\%) and illustrates several key findings.
First, we find that unsurprisingly, accounting for full conversational context is indeed helpful, with even the simple online 
baselines outperforming the 
fixed-window baselines.
On both datasets, CRAFT outperforms all baselines (including the other online models) in terms of accuracy and F1.
Furthermore, although it loses on precision (to CRAFT $-$ CE) and recall (to Cumulative BoW) individually on the Wikipedia data, 
CRAFT has the superior \emph{balance} between 
the two,
 having both a visibly higher precision-recall curve and larger area under the curve (AUPR) than the baselines (Figure \ref{fig:pr_curve}).
This latter property is particularly useful in a practical setting, as it allows moderators to tune model performance to some desired precision without having to sacrifice as much in the way of recall (or vice versa) compared to the baselines and pre-existing solutions.

\section{Analysis}
\label{sec:analysis}
We now examine the behavior of CRAFT in greater detail, to better understand its benefits and limitations. We specifically address the following questions: (1) How much early warning does the 
the model
provide? 
(2) Does the model actually learn an order-sensitive representation of conversational context?\footnote{We choose to focus on the Wikipedia scenario since the conversational prefixes are hand-verified to be civil. For completeness we also report results for Reddit CMV throughout, but they should be taken with an additional grain of salt.}

\begin{figure}
    \centering
    \includegraphics[width=\linewidth]{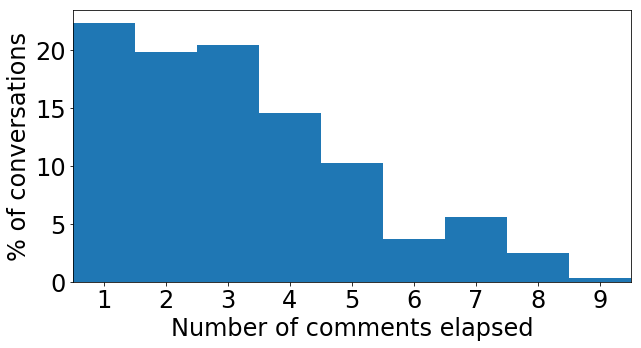}
    \caption{Distribution of number of comments elapsed between the model's first warning and the attack.}
    \label{fig:index_gaps}
\end{figure}

\noindent \textbf{Early warning, but how early?}
The recent interest in forecasting antisocial behavior has been driven by a desire to provide pre-emptive, actionable warning to moderators.
But does our model trigger early enough for any such practical goals?

For each personal attack correctly forecasted by our model, we 
count the number of comments elapsed between the time the model is first triggered and the attack.
Figure \ref{fig:index_gaps} shows the distribution of these 
counts:
 on average, the model warns of an attack 
 3
  comments before 
it actually happens
   (4 comments for CMV).  
To further evaluate how much time this early warning would give to the moderator, we also consider the difference in timestamps between the 
comment where the model first triggers and the 
comment containing the actual attack.
Over 50\% of conversations get at least 3 hours of advance warning (2 hours for CMV).  Moreover, 39\% of conversations get at least \emph{12 hours} 
of early warning 
before they derail.

\noindent\textbf{Does order matter?}
One motivation behind the design of our model was the intuition that comments in a conversation are not independent events; rather, the order in which they appear matters (e.g., a blunt comment followed by a polite one feels intuitively different from a polite comment followed by a blunt one). 
By design, CRAFT has the capacity to learn an order-sensitive representation of conversational context, but how can we know that this capacity is actually used?  It is conceivable that the model is simply computing an order-insensitive ``bag-of-features''.
Neural network models are notorious for their lack of transparency, precluding an analysis of how \emph{exactly} CRAFT models conversational context.
Nevertheless, through two simple exploratory experiments, 
we seek to show
that it does not completely  ignore comment order.

\begin{figure}
    \centering
    \includegraphics[width=\linewidth]{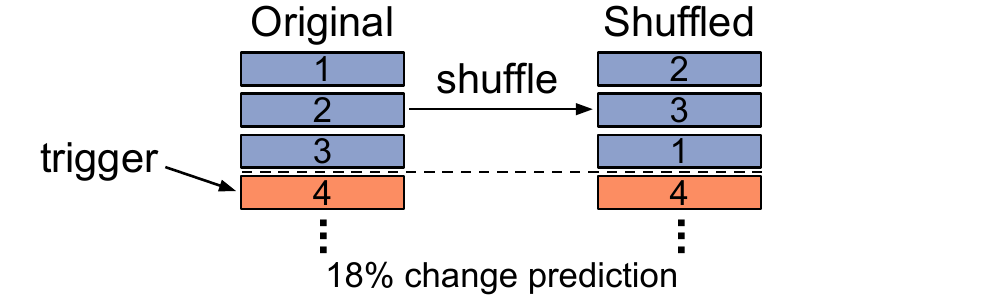}
    \caption{The prefix-shuffling procedure ($t=4$).}
    \label{fig:shuffle}
\end{figure}

The first experiment for testing whether the model accounts for comment order is a \emph{prefix-shuffling} experiment, visualized in Figure \ref{fig:shuffle}.
For each conversation that the model predicts will derail, let $t$ denote the index of the triggering comment, i.e., the index where the model first made a derailment forecast.
We then construct \emph{synthetic} conversations by taking the first $t-1$ comments (henceforth referred to as the \emph{prefix}) and randomizing their order.\footnote{We restrict the experiment to cases where $t \ge 3$, as prefixes consisting of only one comment cannot be reordered.} 
Finally, we count how often the model no longer predicts derailment at index $t$ in the synthetic conversations.
If the model were ignoring comment order, its prediction should remain unchanged (as it remains for the Cumulative BoW baseline), since the actual \emph{content} of the first $t$ comments has not changed (and CRAFT inference is deterministic). 
We instead find that in roughly one fifth of cases (12\% for CMV) the model changes its prediction on the synthetic conversations.
This suggests that CRAFT learns an order-sensitive representation of context, not a mere ``bag-of-features''.

To more concretely quantify how much this order-sensitive context modeling helps with prediction, we can actively prevent the model from learning and exploiting any order-related dynamics.
We achieve this through another type of shuffling experiment, where we go back even further and shuffle the 
comment order in the conversations
 used for pre-training, 
 fine-tuning and testing.
This procedure preserves the model's ability to capture 
signals present
 within the individual comments processed so far, as the utterance encoder is unaffected, but inhibits it from capturing
 any meaningful order-sensitive dynamics.
We find that this hurts the model's performance (65\% accuracy for Wikipedia, 59.5\% for CMV), lowering it to a level similar to that of the  version where we completely disable the context encoder.  

Taken together, these experiments provide evidence that CRAFT 
uses its capacity to
model conversational context in an order-sensitive fashion, and 
that it makes effective use of the dynamics within.
An important avenue for future work would be developing more transparent models that can shed light on exactly \emph{what} kinds of order-related features are being extracted and \emph{how} they are used in prediction.

\section{Conclusions and Future Work}
\label{sec:discussion}
In this work, we introduced a model for forecasting conversational events that processes comments as they happen and takes the full conversational context into account to make an updated prediction at each step.
This model fills a void in the existing literature on conversational forecasting, simultaneously addressing the dual challenges of 
capturing
inter-comment dynamics and 
dealing with an unknown horizon.
We find that our model achieves state-of-the-art performance 
on the task of forecasting derailment
 in two different datasets that we release publicly.
We further show that the resulting system can provide substantial prior notice of derailment, opening up the potential for preemptive interventions by human moderators \cite{seering_shaping_2017}.

While we have focused specifically on the task of forecasting derailment, we view this work as a step towards a more general model for 
real-time
 forecasting of 
other types
  of emergent properties of conversations.
Follow-up work could 
adapt the CRAFT architecture to address other forecasting tasks mentioned in Section \ref{sec:related}---including those for which the outcome is extraneous to the conversation.
We expect different tasks to be informed by different types of inter-comment dynamics, 
and further architecture extensions could add additional supervised fine-tuning 
in order to  direct it to focus on specific dynamics that might be 
 relevant to the task (e.g., exchange of ideas between interlocutors or stonewalling).

With respect to forecasting derailment, 
there remain open questions regarding what human moderators actually desire from an early-warning system, which would affect the design of a practical system based on this work.
For instance, how early does a warning need to be in order for moderators to find it useful?
What is the optimal balance between precision, recall, and false positive rate at which such a system is truly improving moderator productivity rather than wasting their time through false positives?
What are the ethical implications of such a system?
Follow-up work could run a user study of a prototype system with actual moderators to address these questions.

A practical limitation of the current analysis is that it relies on balanced datasets, while derailment is a relatively rare event for which a more restrictive trigger threshold would be appropriate.  While our analysis of the precision-recall curve suggests the system is robust across multiple thresholds ($AUPR=0.7$), additional work is needed to establish whether the recall tradeoff would be acceptable in practice.

Finally, one major limitation of the present work is that it assigns a single label to each conversation: does it derail or not?
In reality, derailment need not spell the end of a conversation; it is possible that a conversation could get back on track, suffer a repeat occurrence of antisocial behavior, or any number of other trajectories.
 It would be exciting to consider
finer-grained forecasting of conversational trajectories, accounting for the natural---and sometimes chaotic---ebb-and-flow of human interactions.

\vspace{0.2in}
\xhdr{Acknowledgements}
We thank Caleb Chiam, Liye Fu, Lillian Lee, Alexandru Niculescu-Mizil, Andrew Wang and Justine Zhang for insightful conversations (with unknown horizon), Aditya Jha for his great help with implementing and running the crowd-sourcing tasks, Thomas Davidson and Claire Liang for exploratory data annotation, as well as the anonymous reviewers for their helpful comments. 
This work is supported in part by the 
NSF CAREER award IIS-1750615 and by the
 NSF Grant SES-1741441.

\bibliographystyle{acl_natbib}
\bibliography{tension-emnlp-autoupdate-jpc,tension-emnlp-autoupdate-C}

\begin{thebibliography}{66}
\expandafter\ifx\csname natexlab\endcsname\relax\def\natexlab#1{#1}\fi

\bibitem[{Abbott et~al.(2011)Abbott, Walker, Anand, Fox~Tree, Bowmani, and
  King}]{abbott_how_2011}
Rob Abbott, Marilyn Walker, Pranav Anand, Jean~E. Fox~Tree, Robeson Bowmani,
  and Joseph King. 2011.
\newblock How {{Can You Say Such Things}}?!?: {{Recognizing Disagreement}} in
  {{Informal Political Argument}}.
\newblock In \emph{Proceedings of the {{Workshop}} on {{Languages}} in {{Social
  Media}}}.

\bibitem[{Akbulut et~al.(2010)Akbulut, Sahin, and
  Eristi}]{akbulut_cyberbullying_2010}
Yavuz Akbulut, Yusuf~Levent Sahin, and Bahadir Eristi. 2010.
\newblock Cyberbullying {{Victimization}} among {{Turkish Online Social Utility
  Members}}.
\newblock \emph{Educational Technology \& Society}, 13(4).

\bibitem[{Allen et~al.(2014)Allen, Carenini, and Ng}]{allen_detecting_2014}
Kelsey Allen, Giuseppe Carenini, and Raymond~T. Ng. 2014.
\newblock Detecting {{Disagreement}} in {{Conversations}} using
  {{Pseudo}}-{{Monologic Rhetorical Structure}}.
\newblock In \emph{Proceedings of {{EMNLP}}}.

\bibitem[{Althoff et~al.(2016)Althoff, Clark, and
  Leskovec}]{althoff_large-scale_2016}
Tim Althoff, Kevin Clark, and Jure Leskovec. 2016.
\newblock Large-scale {{Analysis}} of {{Counseling Conversations}}: {{An
  Application}} of {{Natural Language Processing}} to {{Mental Health}}.
\newblock \emph{Transactions of the Association for Computational Linguistics},
  4.

\bibitem[{Arazy et~al.(2013)Arazy, Yeo, and Nov}]{arazy_stay_2013}
Ofer Arazy, Lisa Yeo, and Oded Nov. 2013.
\newblock Stay on the {{Wikipedia Task}}: {{When Task}}-related {{Disagreements
  Slip Into Personal}} and {{Procedural Conflicts}}.
\newblock \emph{J. Am. Soc. Inf. Sci. Technol.}, 64(8).

\bibitem[{Backstrom et~al.(2013)Backstrom, Kleinberg, Lee, and
  {Danescu-Niculescu-Mizil}}]{backstrom_characterizing_2013}
Lars Backstrom, Jon Kleinberg, Lillian Lee, and Cristian
  {Danescu-Niculescu-Mizil}. 2013.
\newblock Characterizing and {{Curating Conversation Threads}}: {{Expansion}},
  {{Focus}}, {{Volume}}, {{Re}}-entry.
\newblock In \emph{Proceedings of {{WSDM}}}.

\bibitem[{Bahdanau et~al.(2014)Bahdanau, Cho, and
  Bengio}]{bahdanau_neural_2014}
Dzmitry Bahdanau, Kyunghyun Cho, and Yoshua Bengio. 2014.
\newblock Neural {{Machine Translation}} by {{Jointly Learning}} to {{Align}}
  and {{Translate}}.
\newblock In \emph{Proceedings of {{ICLR}}}.

\bibitem[{Cadilhac et~al.(2013)Cadilhac, Asher, Benamara, and
  Lascarides}]{cadilhac_grounding_2013}
Anais Cadilhac, Nicholas Asher, Farah Benamara, and Alex Lascarides. 2013.
\newblock Grounding {{Strategic Conversation}}: {{Using Negotiation Dialogues}}
  to {{Predict Trades}} in a {{Win}}-{{Lose Game}}.
\newblock In \emph{Proceedings of {{EMNLP}}}.

\bibitem[{Cheng et~al.(2017)Cheng, Bernstein, {Danescu-Niculescu-Mizil}, and
  Leskovec}]{cheng_anyone_2017}
Justin Cheng, Michael Bernstein, Cristian {Danescu-Niculescu-Mizil}, and Jure
  Leskovec. 2017.
\newblock Anyone {{Can Become}} a {{Troll}}: {{Causes}} of {{Trolling
  Behavior}} in {{Online Discussions}}.
\newblock In \emph{Proceedings of {{CSCW}}}.

\bibitem[{Cho et~al.(2014)Cho, {van Merrienboer}, Gulcehre, Bahdanau, Bougares,
  Schwenk, and Bengio}]{cho_learning_2014}
Kyunghyun Cho, Bart {van Merrienboer}, Caglar Gulcehre, Dzmitry Bahdanau, Fethi
  Bougares, Holger Schwenk, and Yoshua Bengio. 2014.
\newblock Learning {{Phrase Representations}} using {{RNN
  Encoder}}\textendash{{Decoder}} for {{Statistical Machine Translation}}.
\newblock In \emph{Proceedings of {{EMNLP}}}.

\bibitem[{Collier and Bear(2012)}]{collier_conflict_2012}
Benjamin Collier and Julia Bear. 2012.
\newblock Conflict, {{Criticism}}, or {{Confidence}}: {{An Empirical
  Examination}} of the {{Gender Gap}} in {{Wikipedia Contributions}}.
\newblock In \emph{Proceedings of {{CSCW}}}.

\bibitem[{Curhan and Pentland(2007)}]{curhan_thin_2007}
Jared~R. Curhan and Alex Pentland. 2007.
\newblock Thin {{Slices}} of {{Negotiation}}: {{Predicting Outcomes From
  Conversational Dynamics Within}} the {{First}} 5 {{Minutes}}.
\newblock \emph{Journal of Applied Psychology}, 92.

\bibitem[{Dai and Le(2015)}]{dai_semi-supervised_2015}
Andrew~M. Dai and Quoc~V. Le. 2015.
\newblock Semi-supervised {{Sequence Learning}}.
\newblock In \emph{Proceedings of {{NeurIPS}}}.

\bibitem[{Davidson et~al.(2017)Davidson, Warmsley, Macy, and
  Weber}]{davidson_automated_2017}
Thomas Davidson, Dana Warmsley, Michael Macy, and Ingmar Weber. 2017.
\newblock Automated {{Hate Speech Detection}} and the {{Problem}} of
  {{Offensive Language}}.
\newblock In \emph{Proceedings of {{ICWSM}}}.

\bibitem[{Devlin et~al.(2019)Devlin, Chang, Lee, and
  Toutanova}]{devlin_bert_2019}
Jacob Devlin, Ming-Wei Chang, Kenton Lee, and Kristina Toutanova. 2019.
\newblock {{BERT}}: {{Pre}}-training of {{Deep Bidirectional Transformers}} for
  {{Language Understanding}}.
\newblock In \emph{Proceedings of {{NAACL}}}.

\bibitem[{Duggan(2014)}]{duggan_online_2014}
Maeve Duggan. 2014.
\newblock Online {{Harassment}}.
\newblock http://www.pewinternet.org/2014/10/22/online-harassment/.

\bibitem[{Feldman et~al.(2015)Feldman, Friedler, Moeller, Scheidegger, and
  Venkatasubramanian}]{feldman_certifying_2015}
Michael Feldman, Sorelle~A. Friedler, John Moeller, Carlos Scheidegger, and
  Suresh Venkatasubramanian. 2015.
\newblock Certifying and {{Removing Disparate Impact}}.
\newblock In \emph{Proceedings of {{KDD}}}.

\bibitem[{Galley et~al.(2004)Galley, McKeown, Hirschberg, and
  Shriberg}]{galley_identifying_2004}
Michel Galley, Kathleen McKeown, Julia Hirschberg, and Elizabeth Shriberg.
  2004.
\newblock Identifying {{Agreement}} and {{Disagreement}} in {{Conversational
  Speech}}: {{Use}} of {{Bayesian Networks}} to {{Model Pragmatic
  Dependencies}}.
\newblock In \emph{Proceedings of {{ACL}}}.

\bibitem[{Gao et~al.(2018)Gao, Galley, and Li}]{gao_neural_2018}
Jianfeng Gao, Michel Galley, and Lihong Li. 2018.
\newblock Neural {{Approaches}} to {{Conversational AI}}.
\newblock In \emph{Proceedings of {{SIGIR}}}.

\bibitem[{Garimella et~al.(2017)Garimella, De~Francisci~Morales, Gionis, and
  Mathioudakis}]{garimella_quantifying_2017}
Kiran Garimella, Gianmarco De~Francisci~Morales, Aristides Gionis, and Michael
  Mathioudakis. 2017.
\newblock Quantifying {{Controversy}} in {{Social Media}}.
\newblock \emph{ACM Transactions on Social Computing}, 1(1).

\bibitem[{Girlea et~al.(2016)Girlea, Girju, and
  Amir}]{girlea_psycholinguistic_2016}
Codruta Girlea, Roxana Girju, and Eyal Amir. 2016.
\newblock Psycholinguistic {{Features}} for {{Deceptive Role Detection}} in
  {{Werewolf}}.
\newblock In \emph{Proceedings of {{NAACL}}}.

\bibitem[{Hessel and Lee(2019)}]{hessel_somethings_2019}
Jack Hessel and Lillian Lee. 2019.
\newblock Something's {{Brewing}}! {{Early Prediction}} of
  {{Controversy}}-causing {{Posts}} from {{Discussion Features}}.
\newblock In \emph{Proceedings of {{NAACL}}}.

\bibitem[{Howard and Ruder(2018)}]{howard_universal_2018}
Jeremy Howard and Sebastian Ruder. 2018.
\newblock Universal {{Language Model Fine}}-tuning for {{Text Classification}}.
\newblock In \emph{Proceedings of {{ACL}}}.

\bibitem[{Hua et~al.(2018)Hua, {Danescu-Niculescu-Mizil}, Taraborelli, Thain,
  Sorensen, and Dixon}]{hua_wikiconv_2018}
Yiqing Hua, Cristian {Danescu-Niculescu-Mizil}, Dario Taraborelli, Nithum
  Thain, Jeffery Sorensen, and Lucas Dixon. 2018.
\newblock {{WikiConv}}: {{A Corpus}} of the {{Complete Conversational History}}
  of a {{Large Online Collaborative Community}}.
\newblock In \emph{Proceedings of {{EMNLP}}}.

\bibitem[{Jo et~al.(2018)Jo, Poddar, Jeon, Shen, Ros{\'e}, and
  Neubig}]{jo_attentive_2018}
Yohan Jo, Shivani Poddar, Byungsoo Jeon, Qinlan Shen, Carolyn~P. Ros{\'e}, and
  Graham Neubig. 2018.
\newblock Attentive {{Interaction Model}}: {{Modeling Changes}} in {{View}} in
  {{Argumentation}}.
\newblock In \emph{Proceedings of {{NAACL}}}.

\bibitem[{Jurgens et~al.(2019)Jurgens, Hemphill, and
  Chandrasekharan}]{jurgens_just_2019}
David Jurgens, Libby Hemphill, and Eshwar Chandrasekharan. 2019.
\newblock A {{Just}} and {{Comprehensive Strategy}} for {{Using NLP}} to
  {{Address Online Abuse}}.
\newblock In \emph{Proceedings of {{ACL}}}.

\bibitem[{Kayany(1998)}]{kayany_contexts_1998}
Joseph~M. Kayany. 1998.
\newblock Contexts of uninhibited online behavior: {{Flaming}} in social
  newsgroups on usenet.
\newblock \emph{Journal of the American Society for Information Science},
  49(12).

\bibitem[{Kumar et~al.(2018)Kumar, Hamilton, Leskovec, and
  Jurafsky}]{kumar_community_2018}
Srijan Kumar, William~L. Hamilton, Jure Leskovec, and Dan Jurafsky. 2018.
\newblock Community {{Interaction}} and {{Conflict}} on the {{Web}}.
\newblock In \emph{Proceedings of {{WWW}}}.

\bibitem[{Levitan et~al.(2018)Levitan, Maredia, and
  Hirschberg}]{levitan_linguistic_2018}
Sarah~Ita Levitan, Angel Maredia, and Julia Hirschberg. 2018.
\newblock Linguistic {{Cues}} to {{Deception}} and {{Perceived Deception}} in
  {{Interview Dialogues}}.
\newblock In \emph{Proceedings of {{NAACL}}}.

\bibitem[{Liu et~al.(2018)Liu, Guberman, Hemphill, and
  Culotta}]{liu_forecasting_2018}
Ping Liu, Joshua Guberman, Libby Hemphill, and Aron Culotta. 2018.
\newblock Forecasting the {{Presence}} and {{Intensity}} of {{Hostility}} on
  {{Instagram Using Linguistic}} and {{Social Features}}.
\newblock In \emph{Proceedings of {{ICWSM}}}.

\bibitem[{Luong et~al.(2015)Luong, Pham, and Manning}]{luong_effective_2015}
Minh-Thang Luong, Hieu Pham, and Christopher~D. Manning. 2015.
\newblock Effective {{Approaches}} to {{Attention}}-based {{Neural Machine
  Translation}}.
\newblock In \emph{Proceedings of {{EMNLP}}}.

\bibitem[{McCann et~al.(2017)McCann, Bradbury, Xiong, and
  Socher}]{mccann_learned_2017}
Bryan McCann, James Bradbury, Caiming Xiong, and Richard Socher. 2017.
\newblock Learned in {{Translation}}: {{Contextualized Word Vectors}}.
\newblock In \emph{Proceedings of {{NeurIPS}}}.

\bibitem[{Niculae and
  {Danescu-Niculescu-Mizil}(2016)}]{niculae_conversational_2016}
Vlad Niculae and Cristian {Danescu-Niculescu-Mizil}. 2016.
\newblock Conversational {{Markers}} of {{Constructive Discussions}}.
\newblock In \emph{Proceedings of {{NAACL}}}.

\bibitem[{Niculae et~al.(2015)Niculae, Kumar, {Boyd-Graber}, and
  {Danescu-Niculescu-Mizil}}]{niculae_linguistic_2015}
Vlad Niculae, Srijan Kumar, Jordan {Boyd-Graber}, and Cristian
  {Danescu-Niculescu-Mizil}. 2015.
\newblock Linguistic {{Harbingers}} of {{Betrayal}}: {{A Case Study}} on an
  {{Online Strategy Game}}.
\newblock In \emph{Proceedings of {{ACL}}}.

\bibitem[{Olteanu et~al.(2018)Olteanu, Castillo, Boy, and
  Varshney}]{olteanu_effect_2018}
Alexandra Olteanu, Carlos Castillo, Jeremy Boy, and Kush Varshney. 2018.
\newblock The {{Effect}} of {{Extremist Violence}} on {{Hateful Speech
  Online}}.
\newblock In \emph{Proceedings of {{ICWSM}}}.

\bibitem[{Park et~al.(2018)Park, Shin, and Fung}]{park_reducing_2018}
Ji~Ho Park, Jamin Shin, and Pascale Fung. 2018.
\newblock Reducing {{Gender Bias}} in {{Abusive Language Detection}}.
\newblock In \emph{Proceedings of {{EMNLP}}}.

\bibitem[{Pavlopoulos et~al.(2017)Pavlopoulos, Malakasiotis, and
  Androutsopoulos}]{pavlopoulos_deeper_2017}
John Pavlopoulos, Prodromos Malakasiotis, and Ion Androutsopoulos. 2017.
\newblock Deeper {{Attention}} to {{Abusive User Content Moderation}}.
\newblock In \emph{Proceedings of {{EMNLP}}}.

\bibitem[{{P{\'e}rez-Rosas} et~al.(2016){P{\'e}rez-Rosas}, Abouelenien,
  Mihalcea, Xiao, Linton, and Burzo}]{perez-rosas_verbal_2016}
Ver{\'o}nica {P{\'e}rez-Rosas}, Mohamed Abouelenien, Rada Mihalcea, Yao Xiao,
  C.~J. Linton, and Mihai Burzo. 2016.
\newblock Verbal and {{Nonverbal Clues}} for {{Real}}-life {{Deception
  Detection}}.
\newblock In \emph{Proceedings of {{EMNLP}}}.

\bibitem[{Peters et~al.(2018)Peters, Neumann, Iyyer, Gardner, Clark, Lee, and
  Zettlemoyer}]{peters_deep_2018}
Matthew~E. Peters, Mark Neumann, Mohit Iyyer, Matt Gardner, Christopher Clark,
  Kenton Lee, and Luke Zettlemoyer. 2018.
\newblock Deep {{Contextualized Word Representations}}.
\newblock In \emph{Proceedings of {{NAACL}}}.

\bibitem[{Potash and Rumshisky(2017)}]{potash_towards_2017}
Peter Potash and Anna Rumshisky. 2017.
\newblock Towards {{Debate Automation}}: A {{Recurrent Model}} for {{Predicting
  Debate Winners}}.
\newblock In \emph{Proceedings of {{EMNLP}}}.

\bibitem[{Radford et~al.(2018)Radford, Narasimhan, Salimans, and
  Sutskever}]{radford_improving_2018}
Alec Radford, Karthik Narasimhan, Tim Salimans, and Ilya Sutskever. 2018.
\newblock Improving {{Language Understanding}} by {{Generative Pre}}-training.
\newblock Technical report, {OpenAI}.

\bibitem[{Raskauskas and Stoltz(2007)}]{raskauskas_involvement_2007}
Juliana Raskauskas and Ann~D. Stoltz. 2007.
\newblock Involvement in {{Traditional}} and {{Electronic Bullying Among
  Adolescents}}.
\newblock \emph{Developmental Psychology}, 43(3).

\bibitem[{Ribeiro et~al.(2018)Ribeiro, Calais, Santos, Almeida, and
  Meira~Jr.}]{ribeiro_characterizing_2018}
Manoel~Horta Ribeiro, Pedro~H. Calais, Yuri~A. Santos, Virg{\'i}lio A.~F.
  Almeida, and Wagner Meira~Jr. 2018.
\newblock Characterizing and {{Detecting Hateful Users}} on {{Twitter}}.
\newblock In \emph{Proceedings of {{ICWSM}}}.

\bibitem[{Rosenthal and McKeown(2015)}]{rosenthal_i_2015}
Sara Rosenthal and Kathleen McKeown. 2015.
\newblock I {{Couldn}}'t {{Agree More}}: {{The Role}} of {{Conversational
  Structure}} in {{Agreement}} and {{Disagreement Detection}} in {{Online
  Discussions}}.
\newblock In \emph{Proceedings of {{SIGDIAL}}}.

\bibitem[{Sap et~al.(2019)Sap, Card, Gabriel, Choi, and Smith}]{sap_risk_2019}
Maarten Sap, Dallas Card, Saadia Gabriel, Yejin Choi, and Noah~A. Smith. 2019.
\newblock The {{Risk}} of {{Racial Bias}} in {{Hate Speech Detection}}.
\newblock In \emph{Proceedings of {{ACL}}}.

\bibitem[{Seering et~al.(2017)Seering, Kraut, and
  Dabbish}]{seering_shaping_2017}
Joseph Seering, Robert Kraut, and Laura Dabbish. 2017.
\newblock Shaping {{Pro}} and {{Anti}}-{{Social Behavior}} on {{Twitch Through
  Moderation}} and {{Example}}-{{Setting}}.
\newblock In \emph{Proceedings of {{CSCW}}}.

\bibitem[{Serban et~al.(2016)Serban, Sordoni, Bengio, Courville, and
  Pineau}]{serban_building_2016}
Iulian~V. Serban, Alessandro Sordoni, Yoshua Bengio, Aaron Courville, and
  Joelle Pineau. 2016.
\newblock Building {{End}}-{{To}}-{{End Dialogue Systems Using Generative
  Hierarchical Neural Network Models}}.
\newblock In \emph{Proceedings of {{AAAI}}}.

\bibitem[{Serban et~al.(2017)Serban, Sordoni, Lowe, Charlin, Pineau, Courville,
  and Bengio}]{serban_hierarchical_2017}
Iulian~Vlad Serban, Alessandro Sordoni, Ryan Lowe, Laurent Charlin, Joelle
  Pineau, Aaron Courville, and Yoshua Bengio. 2017.
\newblock A {{Hierarchical Latent Variable Encoder}}-{{Decoder Model}} for
  {{Generating Dialogues}}.
\newblock In \emph{Proceedings of {{AAAI}}}.

\bibitem[{Singh et~al.(2017)Singh, Radford, Huang, and
  Furrer}]{singh_they_2017}
Vivek~K. Singh, Marie~L. Radford, Qianjia Huang, and Susan Furrer. 2017.
\newblock "{{They}} basically like destroyed the school one day": {{On Newer
  App Features}} and {{Cyberbullying}} in {{Schools}}.
\newblock In \emph{Proceedings of {{CSCW}}}.

\bibitem[{Sordoni et~al.(2015{\natexlab{a}})Sordoni, Bengio, Vahabi, Lioma,
  Simonsen, and Nie}]{sordoni_hierarchical_2015}
Alessandro Sordoni, Yoshua Bengio, Hossein Vahabi, Christina Lioma, Jakob~Grue
  Simonsen, and Jian-Yun Nie. 2015{\natexlab{a}}.
\newblock A {{Hierarchical Recurrent Encoder}}-{{Decoder}} for {{Generative
  Context}}-{{Aware Query Suggestion}}.
\newblock In \emph{Proceedings of {{CIKM}}}.

\bibitem[{Sordoni et~al.(2015{\natexlab{b}})Sordoni, Galley, Auli, Brockett,
  Ji, Mitchell, Nie, Gao, and Dolan}]{sordoni_neural_2015}
Alessandro Sordoni, Michel Galley, Michael Auli, Chris Brockett, Yangfeng Ji,
  Margaret Mitchell, Jian-Yun Nie, Jianfeng Gao, and Bill Dolan.
  2015{\natexlab{b}}.
\newblock A {{Neural Network Approach}} to {{Context}}-{{Sensitive Generation}}
  of {{Conversational Responses}}.
\newblock In \emph{Proceedings of {{NAACL}}}.

\bibitem[{Sutskever et~al.(2014)Sutskever, Vinyals, and
  Le}]{sutskever_sequence_2014}
Ilya Sutskever, Oriol Vinyals, and Quoc~V. Le. 2014.
\newblock Sequence to {{Sequence Learning}} with {{Neural Networks}}.
\newblock In \emph{Proceedings of {{NeurIPS}}}.

\bibitem[{Tan et~al.(2016)Tan, Niculae, {Danescu-Niculescu}, and
  Lee}]{tan_winning_2016}
Chenhao Tan, Vlad Niculae, Cristian {Danescu-Niculescu}, and Lillian Lee. 2016.
\newblock Winning {{Arguments}}: {{Interaction Dynamics}} and {{Persuasion
  Strategies}} in {{Good}}-faith {{Online Discussions}}.
\newblock In \emph{Proceedings of {{WWW}}}.

\bibitem[{Vitak et~al.(2017)Vitak, Chadha, Steiner, and
  Ashktorab}]{vitak_identifying_2017}
Jessica Vitak, Kalyani Chadha, Linda Steiner, and Zahra Ashktorab. 2017.
\newblock Identifying {{Women}}'s {{Experiences With}} and {{Strategies}} for
  {{Mitigating Negative Effects}} of {{Online Harassment}}.
\newblock In \emph{Proceedings of {{CSCW}}}.

\bibitem[{Volkova and Bell(2017)}]{volkova_identifying_2017}
Svitlana Volkova and Eric Bell. 2017.
\newblock Identifying {{Effective Signals}} to {{Predict Deleted}} and
  {{Suspended Accounts}} on {{Twitter}} across {{Languages}}.
\newblock In \emph{Proceedings of {{ICWSM}}}.

\bibitem[{Wachsmuth et~al.(2018)Wachsmuth, Syed, and
  Stein}]{wachsmuth_retrieval_2018}
Henning Wachsmuth, Shahbaz Syed, and Benno Stein. 2018.
\newblock Retrieval of the {{Best Counterargument}} without {{Prior Topic
  Knowledge}}.
\newblock In \emph{Proceedings of {{ACL}}}.

\bibitem[{Wang et~al.(2017)Wang, Beauchamp, Shugars, and
  Qin}]{wang_winning_2017}
Lu~Wang, Nick Beauchamp, Sarah Shugars, and Kechen Qin. 2017.
\newblock Winning on the {{Merits}}: {{The Joint Effects}} of {{Content}} and
  {{Style}} on {{Debate Outcomes}}.
\newblock \emph{Transactions of the Association for Computational Linguistics},
  5.

\bibitem[{Wang and Cardie(2014)}]{wang_piece_2014}
Lu~Wang and Claire Cardie. 2014.
\newblock A {{Piece}} of {{My Mind}}: {{A Sentiment Analysis Approach}} for
  {{Online Dispute Detection}}.
\newblock In \emph{Proceedings of {{ACL}}}.

\bibitem[{Warner and Hirschberg(2012)}]{warner_detecting_2012}
William Warner and Julia Hirschberg. 2012.
\newblock Detecting {{Hate Speech}} on the {{World Wide Web}}.
\newblock In \emph{Proceedings of the {{Second Workshop}} on {{Language}} in
  {{Social Media}}}.

\bibitem[{Wiegand et~al.(2019)Wiegand, Ruppenhofer, and
  Kleinbauer}]{wiegand_detection_2019}
Michael Wiegand, Josef Ruppenhofer, and Thomas Kleinbauer. 2019.
\newblock Detection of {{Abusive Language}}: The {{Problem}} of {{Biased
  Datasets}}.
\newblock In \emph{Proceedings of {{NAACL}}}.

\bibitem[{Wulczyn et~al.(2017)Wulczyn, Thain, and Dixon}]{wulczyn_ex_2017}
Ellery Wulczyn, Nithum Thain, and Lucas Dixon. 2017.
\newblock Ex {{Machina}}: {{Personal Attacks Seen}} at {{Scale}}.
\newblock In \emph{Proceedings of {{WWW}}}.

\bibitem[{Yang et~al.(2019)Yang, Chen, Yang, Jurafsky, and
  Hovy}]{yang_lets_2019}
Diyi Yang, Jiaao Chen, Zichao Yang, Dan Jurafsky, and Eduard Hovy. 2019.
\newblock Let's {{Make Your Request More Persuasive}}: {{Modeling Persuasive
  Strategies}} via {{Semi}}-{{Supervised Neural Nets}} on {{Crowdfunding
  Platforms}}.
\newblock In \emph{Proceedings of {{NAACL}}}.

\bibitem[{Yin et~al.(2009)Yin, Xue, and Hong}]{yin_detection_2009}
Dawei Yin, Zhenzhen Xue, and Liangjie Hong. 2009.
\newblock Detection of {{Harassment}} on {{Web}} 2.0.
\newblock In \emph{Proceedings of {{CAW2}}.0}.

\bibitem[{Zhang et~al.(2018{\natexlab{a}})Zhang, Chang,
  {Danescu-Niculescu-Mizil}, Dixon, Thain, Hua, and
  Taraborelli}]{zhang_conversations_2018}
Justine Zhang, Jonathan~P. Chang, Cristian {Danescu-Niculescu-Mizil}, Lucas
  Dixon, Nithum Thain, Yiqing Hua, and Dario Taraborelli. 2018{\natexlab{a}}.
\newblock Conversations {{Gone Awry}}: {{Detecting Early Signs}} of
  {{Conversational Failure}}.
\newblock In \emph{Proceedings of {{ACL}}}.

\bibitem[{Zhang et~al.(2018{\natexlab{b}})Zhang, {Danescu-Niculescu-Mizil},
  Sauper, and Taylor}]{zhang_characterizing_2018}
Justine Zhang, Cristian {Danescu-Niculescu-Mizil}, Christina Sauper, and
  Sean~J. Taylor. 2018{\natexlab{b}}.
\newblock Characterizing {{Online Public Discussions Through Patterns}} of
  {{Participant Interactions}}.
\newblock In \emph{Proceedings of {{CSCW}}}.

\bibitem[{Zhang et~al.(2016)Zhang, Kumar, Ravi, and
  {Danescu-Niculescu-Mizil}}]{zhang_conversational_2016}
Justine Zhang, Ravi Kumar, Sujith Ravi, and Cristian {Danescu-Niculescu-Mizil}.
  2016.
\newblock Conversational {{Flow}} in {{Oxford}}-style {{Debates}}.
\newblock In \emph{Proceedings of {{NAACL}}}.

\end{thebibliography}
\end{document}